# SUPPORTING LANGUAGE LEARNERS WITH THE MEANINGS OF CLOSED CLASS ITEMS

Hayat Alrefaie [1] and Prof. Allan Ramsay [2]

[1]The University of Manchester, Manchester, United Kingdom
alrefaih@cs.man.ac.uk

[2]The University of Manchester, Manchester, United Kingdom
allan.ramsay@manchester.ac.uk

## ABSTRACT

*The process of language learning involves the mastery of countless tasks: making the constituent sounds of the language being learned, learning the grammatical patterns, and acquiring the requisite vocabulary for reception and production.*

*While a plethora of computational tools exist to facilitate the first and second of these tasks, a number of challenges arise with respect to enabling the third. This paper describes a tool that has been designed to support language learners with the challenge of understanding the use of 'closed-class' lexical items.*

*The process of learning the Arabic for 'office' is مكتب (mktb) is relatively simple and should be possible by means of simple repetition of the word. However, it is much more difficult to learn and correctly use the Arabic equivalent of the word 'on'. The current paper describes a mechanism for the delivery of diagnostic information regarding specific lexical examples, with the aim of clearly demonstrating why a particular translation of a given closed-class item may be appropriate in certain situations but not others, thereby helping learners to understand and use the term correctly.*

## 1. INTRODUCTION

This study seeks to assist learners of Arabic to correctly understand and utilise closed-class items. The particular focus of this study is to support learners in the correct use of spatial prepositions. The process of learning vocabulary is often extremely labour intensive, as fluency in a language relies upon the acquisition of an extremely comprehensive individual lexicon (for example, an adult speaker will typically have active knowledge of 10-40K words). Although the acquisition of these words may require substantial investment of time and effort, it often involves the memorisation of large numbers of pairs and can therefore be as straightforward as learning that the equivalent of '*office*' is مكتب (*mktb*), the equivalent of '*man*' is رجل (*rgl*), or the equivalent of '*boy*' is ولد (*wld*).

Closed-class items constitute a different type of challenge. By definition, closed-class items are relatively uncommon, with English and Arabic each having a few dozen prepositions. However, their meanings are often somewhat fluid and so choosing a direct equivalent for one preposition can be difficult, if not impossible.

Typically, in order to learn the meaning of a word, common practice is to check the meaning in a dictionary. However, this is often ineffective for words like '*in*', which is defined as a word

'used to indicate location or position within something'. Checking the word '*within*' in the same dictionary explains that it means 'inside something', and then the definition of '*inside*' is provided as 'an interior or internal part or place: the part within'. These circular definitions are often unhelpful for language learners and fail to offer clear advice on functional usage, such as whether '*in*' or '*on*' is correct in a particular case (why do people get in a car but on a bus? Why is my office in the computer science building but on the second floor? . . . ).

Understanding the function of prepositions work requires the knowledge that they express a relationship between some parts of some facets of the entities being related. The latter half of this idea is informed by Pustejovsky's [30] notion of 'qualia structure' or Dowty's [15] more general notion of 'logical space':

There will be as many axes of logical space as there are kinds of measurement. [ . . . ] *Each axis might have a different mathematical structure according to the discriminations that can be appropriately made in each case* [p. 126].

Consider the examples in (1):

(1)    a. I read it in the park.
        b. I read it in the Guardian.
        c. I read it in January.

Each of these examples describes a relationship between a reading event and another thing through the use of the same preposition: (1a) relates the reading event to a place; (1b) expresses a relationship between the reading event and a source of information; and (1c) links the reading event with an interval. It can be difficult to understand that an entity can be linked to such different objects through the same relationship. Therefore, an assumption can be made with respect to the varied ways in which these objects can be perceived, so the preposition begins by determining compatible views of the figure and ground. This allows (1a) to be concerned with the physical act of looking at an object and using received light to create a picture of that object on your retina; (1b) to describe the abstract transferral of information from an external source into your mind; and (1c) to focus on the temporal view, with something that begins at one point in time and ends at a later point in time.

Unlike Pustejovsky, this study makes no claim about the specific number of potential views in natural language. Instead, the notion is implemented through the use of a simple and efficient semi-partitioned type lattice. We will write $T_1 \sim T_2$ to state that $T_1$ and $T_2$ are compatible with regards to the type lattice.

Once compatible views of the figure and ground have been found, the preposition then provides information on the relationship between given views of the two entities being related. For example, the word '*in*' states that the (selected view of the) ground has an interior which encloses some region in the relevant space, and that every part of the (compatible view of the) figure is contained in this region. In contrast, the word '*on*' says that the (selected view of the) ground has an orientable surface, and that the (compatible view of the) figure has a set of contiguous points which lie an arbitrarily small distance from the surface.

Reducing '*F in G*' to $(S \sim S')\&(view(F, S) \subseteq interior(view(G, S')))$ avoids the circularity of saying that '*in*' means '*inside*' because we can give precise definitions of the interior of some entity in a given space, e.g.

- If *X* has an embedding *E* into $R^n$, then interior*(X)* is the set of members p of *X* such that *E(p)* is not a member of *X* but there are members $p_1$, $p_2$ of X such that *E(p)* lies on the line from $E(p_1)$ to $E(p_2)$. Insofar as our everyday notion of physical space can be captured by mapping points in space to $R^3$, this enables us to talk about the interiors of physical objects.

- If *X* is a dense partially ordered set of points, an interval has a set of lower bounds (points which are not preceded by any other point in the set) and upper bounds (defined likewise). The interior of an interval is then the interior of the set consisting of its lower and upper bounds. Most of the commonsense approaches to time discussed by authors such as [32]; [33]; [4] and linguistic approaches discussed by authors such as [35]; [23] involve at least dense partial orderings, and hence can be covered by this definition.

- The interior of a simple unstructured set is the set itself. To say that someone is in my class is simply to say they are a member of the set of people who I am teaching.

In other words, if we have two items placed in some abstract space with an appropriate structure (distance if we want to use '*at*', regionhood if we want to use '*in*', vertical direction if we want to use '*on*', . . .) then we can locate them with respect to each other in that space, using clear definitions that exploit the structure of the space. Of course, the two items have to exist in the same space–it is not possible to say that a physical object is contained in a temporal space (* '*London is in January*'), or that a temporal object is contained in an abstract space (* '*January is in my best paper on syntax*'). Hence the first task when interpreting a preposition is to find views of the entities in question which are in the same space.

Not every item in a given space, however, has the properties required by a given preposition. It is perfectly possible to have a physical object which does not have a non-empty interior (* '*my office is in the second floor*'), or an interval that does not have upper and lower bounds (* '*I have been knowing how to play tennis since I was 12*' is problematic because once you know something, you don't stop knowing it, so the interval denoted by the verb '*know*' has no (natural) upper bound). For a preposition to locate one entity with respect to another, the entities in question have to have the right properties as well as being of the right general kinds. We will return to this in Section 2.

## 1.1. English and Arabic prepositions

The main aim of the current paper is to provide English learners of Arabic with diagnostic information about their use of essentially spatial prepositions by exploiting the notions described above. The problem is that while a given English preposition may have a 'typical' translation into Arabic, there will be numerous exceptions to this standard translation, and the causes of these exceptions are hard to discern.

Consider (2), where في(*fy*) and على(*ElY*) are normally taken mean '*in*' and '*on*' respectively.

(2)  a. My office is on the second floor.
 b. *مكتبي على الطابق الثاني.(*mktby Ely AlTAbq AlvAny.*)
 c. مكتبي في الطابق الثاني(*mktby fy AlTAbq AlvAny.*)

The natural way for an English person to think of a floor is as a flat surface, with no interior, so that if you want to locate a room with respect to a floor you have to say that the room is '*on*' the floor. Arabic speakers, on the other hand, conceptualize a طابق *(TAbq)* as a container, like a very large room, and hence want to say the مكتب *(mktb)* is inside the طابق *(TAbq)* rather than on its surface. Why speakers of one language have different ideas of what various everyday objects are like is beyond the scope of the work reported here. It just seems to be the case. What matters is that it does seem to be the case, and it is this underlying conceptual framework that drives the choice of particular prepositions for linking different kinds of figures and grounds.

## 2. MEANINGS, MODELS AND INFERENCE

To make any computational sense of this idea, we have to give precise characterisations of how the various prepositions work, and we have to describe the entities that they can link. We use a framework in which natural language texts are translated, through compositional rules, into some suitable logic; we then write meaning postulates to capture the consequences of using particular combinations of words; and finally we carry out abductive inference to investigate the use of different prepositions for linking the same figure and ground in two languages.

### 2.1 Logical forms

The framework used here follows the tradition started by Montague [26]; [16] of annotating grammatical rules with semantic interpretations which can be used to 'construct an interpretation by combining the interpretations of the parts'. This process lets you construct formal paraphrases ('logical forms', LFs) in some suitable logic of any sentences that you can parse. There have been many developments in this area since Montague's original presentation of this idea: in particular, the choice of a suitable logic has been extended to cover the notion that natural language is situated, i.e. that utterances have to be interpreted with respect to the context in which they are produced [21]; [6]; [19]; [18] and to take account of the fact that natural languages make extensive use of constructions such as defaults [34]; [5] and propositional attitudes which require highly expressive logics [37]; [1]; [2]. The general principle remains the same: if you have a structural analysis of a text which is obtained by applying the rules of a grammar, you can annotate the rules of that grammar with semantic composition rules which will let you produce an LF. And if you have an LF in some suitable logic, and you have an inference engine for that logic, then you can reason about what has been said.

There are, clearly, two limiting factors to be considered here:

**The structural analysis has to be obtained by using the rules of a grammar** Modern data driven parsers often have no explicit grammar [25]; [28]; [27], or have a very (very) large context-free grammar induced from a treebank [11]; [9]. Clearly, it is not possible to annotate the rules of the grammar being used if there is in fact no grammar; and it is infeasible to do so in cases where the induced grammar contains tens of thousands of rules. Bos?Bos:08a describes a system which produces semantic interpretations using a fairly wide-coverage grammar, but grammar-based parsers can become very slow when faced with long (40+ words) sentences. In the present context, however, we are working with comparatively short sentences, since the task is to compare interpretations of English sentences set as translation exercises with interpretations of putative translations into Arabic. As such we can control the length of the sentences under consideration, thus making it possible to use an HPSG-like grammar and a fairly orthodox chart parser. The grammar we are using has an underlying set of principles, together with a small set of language specific constraints. We can thus use essentially the same grammar for parsing the English and Arabic texts. This has the enormous advantage that exactly the same machinery can be used for constructing LFs for both languages, which makes it comparatively easy to compare them. Fig 1 shows the LFs for (2a) and (2b).

These LFs are fairly orthodox. The main points to note are

- The LFs contain the surface utterance type–*claim, query*, . . . There are two reasons for this: (i) the utterance type is part of the meaning of the utterance, and hence it naturally belongs in the LF; and (ii) in some cases, notably the English determiner *'any'*, there are quantifiers which are naturally seen as out scoping the utterance type: these simply cannot be accommodated without including the utterance type in the LF.
- Referring terms in the original text are represented as reference terms of the form *ref(λX (. . . X . . .))*. Such terms have to be anchored with respect to the context in order to determine the significance of the utterance[6].
- The terms in the Arabic LF make use of predicates with names like كتب *office*, where we use the English gloss to distinguish between distinct terms with the same root (the stem مكتب(*mktb*) is derived from the root كتب(*ktb*), so since we use roots rather than stems in our lexicon it is the root that appears in the LF).

*utt*(*claim*,
   *on*(*ref*(λ*Eown*(*ref*(λ *F*(*speaker*(*F*))), *E*)&*office*(*E*)),
      *ref*(λ*Gfloor*(*G*)&*second*(*G*, λ*H*(*floor*(*H*))))))

*utt*(*claim*,
   *ElY*(*ref*(λ*Downer*(*ref*(λ *E*(*speaker*(*E*))),*D*)&كتب *office*(*D*)),
      *ref*(λ*F* طابق *floor*(*F*) & ☐☐☐ *second*(*F*))))

Figure 1. Logical forms for (2)

**You have to have an inference engine for the chosen logic** There is no point in constructing any kind of meaning representation unless you are going to **do** something with it. In general, what you want to do when confronted with an utterance is to reason about it: to see what follows from it, to think about why the speaker produced, to try to decide what you should do in response to it, . . .

This means that you have to have some kind of engine that is capable of carrying out inference over formulae in the formalism that you have chosen as your representation. This is true no matter whether the meaning representation is just the original string of words and the inference engine is string-edit distance between sequences of words; or the meaning representation is the dependency tree for the text and the inference engine is some form of tree-edit distance [39]; [3], as is commonly employed for textual entailment (Dagan et al., 2005); or whether it is a translation into first-order logic and the inference engine is a theorem prover for first order logic [7]. Simple string/tree-based representations have two advantages: (i) they are robust, since to construct the representation you simply invoke a morphological analyser (for string-based representation) or some kind of dependency parser (for tree-based ones). It doesn't even matter that the analyser may not be particularly accurate, since you can get some information out of a parse tree even if contains some mistakes. (ii) the inference engines are fairly easy to implement–string edit distance algorithms which exploit lexical relations when decided the cost of substituting one term for another are straightforward and the more complex ones (such as [39]) are widely downloadable. They do not, however, support detailed chains of reasoning, and they do not generally pay much attention to whether the reasoning that is carried out is sound. Indeed, typical measures of success for shallow inference report on precision and recall values, where precision means 'How often are the entailments discovered by the system valid?' In other words, such systems recognise from the outset that the inference that they are carrying out may not be sound. Approaches based on translation into logic have the obverse advantages: (i) they are sound, since the target logics have well-defined proof and model theories, so the steps that can be carried out are guaranteed to lead to true conclusions if the premises are true. (ii) They are efficient, and hence can be used to carry out long chains of inference. They also have the obverse problems. In particular, it is difficult to construct them, since the standard approach involves parsing the input text using a grammar which has been annotated with semantic interpretations, and as noted above most systems for parsing free text with long sentences use either a grammar which has been inferred from a treebank, and which is very hard to annotate, or no grammar at all. For the current task, the need to carry out detailed chains of inference outweighs the difficulties that occur when you try to parse long natural texts. For our particular task we have a great deal of control over what the learner will write, because their task is to choose the right prepositions to translate a set of target texts. If the target texts are fairly simple (say up to ten or twelve words) then the texts that users generate will also be fairly simple–you would hardly translate '*I am on the train*' by a sentence with 40 words in it. We therefore use translations into property theory[37]; [10] ,which allows properties and propositions to be treated as arguments and hence allows for finer-grained distinctions than are permitted in first-order logic, as our target representations; and we use a version of the theorem prover described in [31] as our inference engine.

## 2.2 Meaning postulates

In order to reason about the consequences of some utterance, you have to have some knowledge to reason with. If you have a perfect translation from English or Arabic into some formalism and you have a theorem prover for that formalism, you will still not be able to do anything unless you also have an appropriate body of background knowledge.

The knowledge that we require involves describing the circumstances under which some preposition can use for locating one entity with respect to another. We write these down as meaning postulates in property theory–typical rules for describing the use of the preposition '*in*' and '*on*' are given in Fig 2.

$\forall B \forall C$: {$in(B, C)$}
    $\forall D \forall E \forall F$: {$view(B, F)$ & $type(F,D)$ & $dim(F, E)$}
        $\forall G \forall H$: {$view(C,H)$
            & $type(H,G)$ & $(D \sim G)$ & $dim(H, E)$}
            $\forall I$: {$interior(H, I)$}
                $F \subset I$ & $located(B, C)$

$\forall B \forall C$: {$on(B, C)$}
    $\forall D \forall E$ : {$view(B, E)$ & $type(E,D)$}
        $\forall F \forall C$: {$view(C,G)$ & $type(G, F)$ & $D \sim F$}
            $\forall H$: {$bottom(E,H)$}
                $\forall I$: {$surface(G, I)$}
                      $touching(H, I)$ & $located(B, C)$

Figure 2: Meaning postulates for '*in*' and '*on*'

The first rule in Fig 2 says that if the preposition '*in*' has been used to link to entities *B* and *C* then B will be successfully located with respect to *C* if there are views *F* and *H* of *B* and *C* where *F* and *H* have compatible types *D* and *G* and *F* has an interior *I* with the same dimensionality *E* as *H*. Under those circumstances *F* is a subset of *I,* and we have located *B* with respect to *C*. In other words '*in*' works for locating a figure entity with respect to a ground if the ground has a suitable interior. Similarly, the second rule in Fig 2 says that the preposition *'on'* may be used to locate a figure with respect to a ground if the ground has a view which has a surface, and the figure has to have a compatible view which has a bottom.

Given that Arabic has prepositions, في(*fy*) and على (*ElY*), which seem to carry the same meanings as 'in' and 'on', we can write parallel meaning postulates for them. The Arabic spatial prepositions seem to mean almost exactly the same as the English ones, and hence we use exactly the same rules for them. We also need meaning postulates to tell us what views, with what properties, particular words induce. The two languages do differ at this point: to continue with our running example, 'building' and مبنى(*mbnY*) both denote entities with interiors; but although *'floor'* and طابق (*TAbq*) are roughly equivalent, a floor is a 2D surface, with no interior, whereas a طابق is a 3D space with an interior and no surface. To capture these observations, we need some general rules about what kinds of entities have interiors, and surfaces, and so on

$\forall B \forall C \forall D: \{embedding(C, B, R^D)\}$
    $\exists E interior(B, E) \& \exists F(embedding(F, E, R^D))$

$\forall B \forall C: \{embedding(C, B, R^3)\}$
    $\forall D : \{type(B,D)\} \& ( \exists E top(B,E) \& type(E,D)) \& \exists F bottom(B, F) \& type(F,D)$

$\forall B \forall C \forall D: \{embedding(C, B, R^2) \& orientable(C)\}$
    $\exists E(surface(B, E))$

. . .

Figure 3: Rules about $R^N$

The first rule in Fig. 3 says that if there is an embedding *C* of B into $R^N$ then *B* has an interior which also has an embedding into this set. This is not the only kind of entity that can have an interior, but it is a very common one. The second rule says that if there is an embedding of *B* into $R^N$ then *B* has members which are top and bottom points. Again, there are other kinds of entity that can have top and bottom points, but this is a common one. The third says that if there is an orientable embedding of *B* in to $R^2$ then B has a surface. The rules in Fig. 3 are themselves derivable from basic properties of RN and the definitions of *interior*, *top* and *bottom* for these spaces. We are not, in this paper, concerned with deriving such rules from first principles (see [29] for work along these lines). What we need is to be able to say what kinds of things have interiors and so on. We then need to say things about particular entities. It is at this point that the nuances of different languages emerge.

$\forall B :\{office(B)\}$
   $type(B,physical) \& \exists C(embedding(C, B, R^3))$

$\forall B :\{building(B)\}$
   $type(B, physical) \& \exists C(embedding(C, B, R^3))$

$\forall B :\{floor(B)\}$
   $type(B, physical)$
   $\& \exists C embedding(C, B, R^2) \& orientable(C)$

$\forall B :\{كتب\ office(B)\}$
   $type(B, physical) \& \exists C(embedding(C, B, R^3))$

$\forall B :\{ مبنى\ building(B)\}$
   $type(B, physical) \& \exists C(embedding(C, B, R^3))$

$\forall B :\{ طابق\ floor(B)\}$
   $type(B, physical) \& \exists C(embedding(C, B, R^3))$

Figure 4: Buildings, offices & floors

Fig. 4 says that the English and Arabic words for building and office denote entities with the same spatial properties, whereas the words 'floor' and طابق*(TAbq)*, despite being rough translations of one another, differ when you look at them in detail–a 'floor' is an oriented 2D surface, a طابق *(TAbq)* is a 3D container.

Given these descriptions, it is easy to see why English and Arabic speakers use a preposition (*'in'*, ف *(fy))* which talks about interiors for locating an office with respect to a building, but English speakers use one (*'on'*) which talks about surfaces for locating an office with respect to a floor while Arabic speakers continue to useTYPE EQUATION HERE. في(*fy*) for these kinds of objects. *'in'* just won't work for locating the office with respect to the floor, because a floor doesn't have an interior.

# 3 MODELS VS ABDUCTION

Turning this observation into a CALL tool is not straightforward. There are two plausible ways to proceed: we can either use the theorem prover as a model builder, and compare the models that are generated from an English utterance and the English meaning postulate and an Arabic utterance and the Arabic meaning postulates; or we can adapt the theorem prover to work abductively, trying to fill in the gaps in the use of the prepositions.

## 3.1 Model building

Given a set of axioms, a tableau-style theorem prover can be used as a model builder. In order to use a tableau engine to prove that a goal *G* follows from a set of axioms *A* you have to negate the gaol and show that every branch of the resulting tableau is closed. The key point here is that a branch of a tableau consists of a set of literals: if the branch is closed, then it contains some directly contradictory pair of literals, and hence cannot have a model. If every branch is closed, then there is no model for the initial set of propositions. Since the initial set of propositions was $G \cup \{\mathbf{not}(\mathbf{A})\}$, this means that it is not possible for G to be true and A to be false, and hence that A is true whenever G is. Just before the closing literal is added, however, the set of literals that make up a branch are consistent, and hence have a model. Indeed, the set of positive literals in this set **is** a Herbrand model.

Suppose, then, that we apply the machinery of a tableau theorem prover to a consistent set of clauses. If at some point we find that we have an open branch, and that there is nothing further that can be added to that branch. Then again the set of positive literals is a model.

Suppose we take two sentences, one in English and Arabic, and try to build models for them along with an axiomatisation of the background knowledge relating to the terms that they contain. If the two sentences mean the same, then we might expect the models to be equivalent. If they are not, then it is a reasonable supposition that the differences between them reflect different connections between the terms that they contain and the relevant background knowledge. It may then be possible to provide feedback about these differences, and hence to help the learner to see how terms that appear to be mutually equivalent differ when used in different contexts.

To make this concrete, imagine that our learner has been asked to translate 'My office is on the second floor.', and they have done so by producing the sentence. مكتبي على الطابق الثاني*(mktby Ely AlTAbq AlvAny.)* This is not an unreasonable translation, since على (*Ely*) is often taken as the translation of 'on'. Unfortunately, as we have just seen, it is wrong, because the طابق (*TAbq*) is conceptualised as a container rather than a surface. The models for these two sentences are shown in Fig. 5. The two models in Fig. 5 contain a substantial amount of information that is not present in the original LFs for the two sentences. That is as it should be–a model contains information that arises when you combine an interpretation with a body of background knowledge. If you look carefully, it is possible to see that the English model contains the fact that the office, #3228, has been located with respect to the floor, #3229, and that this has happened because the floor has an oriented embedding #3274 into $R^2$ and hence has a surface, #3265, whereas the طابق floor, #3655, in the Arabic model has an embedding into $R^3$ rather than $R^2$, and hence has no surface for the كتب office, # 3654, to rest on. The problem is that you have to look very carefully in order to spot the relevant differences between the two models. Given that we are concerned with the use of prepositions for locating one entity with respect to another, it makes sense to look for the entries in the models that correspond directly to the prepositions, and to then look to see whether the entities referred to in these entries are located with respect to one another.

```
'my office is on the second floor'                مكتبي في الطابق الثاني. (mktby fy ālṭābq ālṯāny.)
[office(#3228),                                   [كتب office(#3654),
bottom(#3228, #3272),                             ElY(#3654, #3655),
dim(#3228, 3),                                    bottom(#3654, #3691),
interior(#3228, #3266),                           interior(#3654, #3685),
located(#3228, #3229),                            type(#3654, physical),
on(#3228, #3229),                                 في floor(#3655),
type(#3228, physical),                            second(#3655),
floor(#3229),                                     bottom(#3655, #3692),
dim(#3229, 2),                                    interior(#3655, #3687),
interior(#3229, #3268),                           type(#3655, physical),
second(#3229, λB(floor(B))),                      interior(#3685, #3686),
surface(#3229, #3265),                            interior(#3687, #3688),
type(#3229, physical),                            embedding(#3689, #3685, R³),
dim(#3266, 3),                                    embedding(#3690, #3687, R³),
interior(#3266, #3267),                           type(#3691, physical),
dim(#3268, 2),                                    type(#3692, physical),
interior(#3268, #3269),                           embedding(#3693, #3654, R³),
embedding(#3270, #3266, R³),                      embedding(#3694, #3655, R³),
embedding(#3271, #3268, R²),                      owner(#user, #3654),
member(#3272, #3228),                             type(#user, human)]
touching(#3272, #3265),
type(#3272, physical),
embedding(#3273, #3228, R³),
orientable(#3274),
embedding(#3274, #3229, R²),
own(#user, #3228)]
```

Figure 5: Models for (2a) and (2b)

### 3.2 Abductive reasoning

There are, however, some problems associated with using comparison between models as a source of feedback. Firstly, you have to be able to link items in the two models, despite the fact that they are described using different terms (drawn, as they are, from source texts in different languages). This requires a set of bilingual equivalents, so that an item described as a floor in one model can be matched to an item described as a طابق in the other. This can be tricky, since the conditions under which a term can be given a particular translation are complex (if they were not, then they would not cause problems for learners). Secondly, a single consistent set of formulae can have multiple models. There is no easy guarantee that the first model that is produced for one sentence is the best match for some model of the second. If the English sentence has models $M^E_1, M^E_2, M^E_3, ...$ and the Arabic sentence has models $M^A_1, M^A_2, M^A_3, ...$ then it may be that the best match is between $M^E_5$ and $M^A_7$. But trying to compare every model of the English with every model of the Arabic to find out which ones have the smallest number of mismatches is going to be very time-consuming.

We therefore also plan to investigate the use of 'abductive' inference. Suppose that our learner has written a sentence using the preposition على as a translation of the English preposition *'on'*. We have a collection of rules that describe the conditions under which some preposition **will** successfully locate the figure with respect to the ground, as in Fig. 2. Given a model, we can look for places where someone has used a preposition without locating the relevant entities by looking at each rule that has *located(B,C)* has its consequent where the **first** element of the antecedent (which will always specify that some preposition has been used) is satisfied but the consequent is not. We can then use the theorem prover abductively, by allowing proofs with missing steps, where we can prespecify the number of steps that can be omitted. In order to find the minimal set of missing steps, we call the theorem prover once allowing one missing step, then allowing two, then . . . until a proof is found. This does increase the time taken for finding a suitable proof, but if we simply allow proofs with unbounded numbers of missing steps then we are likely to be led to incorrect diagnoses. Thus if we try to carry out an abductive proof that two items mentioned in the model of (2b) are located, we will find that the rule using على (*Ely*) would have worked if we had been able to prove *surface(#3655, B)* for some *B*. So the primary reason why (2b) failed to locate the مكتب (*mktb*) is that the طابق (*TAbq*) does not have a surface. We could, therefore, provide a message to that effect to the learner as the first round of feedback–'You tried to use على (*Ely*) as the translation of *'on'*, but it doesn't work in this case because although طابق (*TAbq*) is the correct translation of *'floor'*, a طابق (*TAbq*) does not have a surface'. We can follow this up with more detailed feedback. If the learner is not satisfied with this as a comment, he or she can ask why a طابق (*TAbq*) does not have a surface. An abductive proof of *surface (#3655, B)* produces the response that this would have worked if we had been able to show *embedding (B,#3655,R2)&orientable(B)*. It is harder to see how this could be turned into a comprehensible diagnostic message for the learner, but the general principle is simple. The reason why some attempted translation fails to produce a target consequence can be found by carrying an abductive proof of the target and recording the missing steps. In many cases, these missing steps can be converted to meaningful diagnostic messages; and if the learner wants to know more about what went wrong, then abductive proofs of the missing steps themselves can be derived, though as this process gets deeper the rules that are being used become more and more abstract, to the point where they will not in fact produce output will help the learner to understand what they have done wrong.

## 4 FUTURE WORKS

The transformation of the work described here into a functional CALL tool requires the completion of two principal tasks:

1. Meaning postulates need to be written for a considerably wider set of open-class words. Rules have already been prepared for the major special prepositions (Fig. 2), as well as for general uses like the interior, surface or ceiling of an object (Fig. 3). In order to make the tool suitable for use within the classroom context, however, it will be necessary to prepare a large set of entities able to enter into these relationships, as illustrated by the examples in Fig. 4.

2. The output of the abductive reasoner, which constitutes a set of missing steps, must be converted into useful, comprehensible diagnostic messages. At this time, it is expected that canned phrases with slots in will be suitable for this function. However, the missing steps contain Skolem constants, which will be unknown to language learners, meaning that care must be taken when using these steps to derive suitable referring expressions [13]; [38].

The first task, writing a larger set of open-class words, will be time consuming but relatively simple. The second is likely to be more challenging, although we believe that the outcome should offer richer and more useful information about the use of particular prepositions than is currently available through the use of statistical approaches, such as those proposed by [14]; [17]; [36].

**Authors**

**1. Hayat M A. Alrefaie**

Postgraduate Student
School of Computer Science
University of Manchester

**2. Prof Allan Ramsay**

Professor of Formal Linguistics
School of Computer Science
University of Manchester